\documentclass{article}
\usepackage{iclr2027_conference,times}

\usepackage{amsmath,amssymb,amsthm,booktabs,colortbl,graphicx,microtype,multirow}
\usepackage[
  colorlinks=true,
  linkcolor=blue,
  citecolor=cyan,
  urlcolor=magenta
]{hyperref}
\usepackage{url}
\usepackage{xspace}

\newcommand{\method}{DoppelVet}
\newcommand{\readout}{\rho}
\newcommand{\outcome}{Y}
\newcommand{\trace}{\tau}
\newcommand{\updateSpace}{\mathcal{U}}
\newcommand{\states}{\mathcal{S}}

\definecolor{tableheader}{RGB}{228,235,244}
\definecolor{tablebest}{RGB}{224,242,231}
\definecolor{tablegroup}{RGB}{244,247,250}
\definecolor{tablemean}{RGB}{236,246,239}
\newtheorem{proposition}{Proposition}

\title{DoppelVet: Comparative Safety Verification for Self-Evolving Agents}
\author{%
\begin{tabular}{@{}c@{\hspace{2.4em}}c@{\hspace{2.4em}}c@{}}
\textbf{Yifan Duan} & \textbf{Du Yin} & \textbf{Siyuan Zheng}\\[0.35em]
\textbf{Chao Xue} & \textbf{Lihuan Li} & \textbf{Flora D. Salim}\thanks{Corresponding author.}
\end{tabular}\\[0.7em]
UNSW Sydney, Australia\\[0.25em]
{\small\texttt{yifan.duan3@unsw.edu.au}\quad\texttt{flora.salim@unsw.edu.au}}
}

\makeatletter
\renewcommand{\@maketitle}{%
  \vbox{\hsize\textwidth
    {\LARGE\scshape\@title\par}
    \vskip 18pt
    \hbox to \textwidth{\hfil
      \begin{tabular}{@{}c@{}}\@author\end{tabular}%
      \hfil}%
    \vskip 0.3in minus 0.1in
  }%
}
\makeatother

\hypersetup{
  pdftitle={DoppelVet: Comparative Safety Verification for Self-Evolving Agents},
  pdfauthor={Yifan Duan, Du Yin, Siyuan Zheng, Chao Xue, Lihuan Li, Flora D. Salim},
  pdfsubject={Preprint}
}

\iclrfinalcopy
\begin{document}
\maketitle
\lhead{Preprint. Under review.}

\begin{abstract}
Self-evolving agents improve by retaining feedback as reusable memories or
skills, but every mistaken approval can shape many future behaviors. Two
updates can provide exactly the same evidence to a verifier yet produce
opposite safety outcomes, so the verifier cannot distinguish them from that
evidence alone. We introduce \method{}, a verification framework that checks
proposed updates before they enter an agent's persistent state. Without access
to safety labels, \method{} constructs an alternative that matches every signal
used by the existing verifier. It replays both updates from the same state and
directly compares their behaviors in both presentation orders before deciding
whether to retain the candidate. We evaluate \method{} on \method{}-Grid and
\method{}-Tool, covering executable programs and natural-language memories.
Across nine open-weight and proprietary verifiers, \method{} raises mean
harmful-update detection AUC from 0.762 to 0.940 on \method{}-Grid and from
0.681 to 0.826 on \method{}-Tool, outperforming VaG and AgentDoG 1.5.
On \method{}-Grid, it reduces trajectories experiencing degradation from
30.39\% to 0.43\% over a ten-generation self-update loop and leaves no final
degradation.
\end{abstract}

\section{Introduction}

Large language models increasingly serve as agents that plan, use tools, and
act across extended interactions \citep{yao2023react,liu2024agentbench}. These
agents navigate digital environments,
explore embodied worlds, and complete multi-step workflows involving external
tools and users
\citep{zhou2024lats,wang2023voyager,zhou2025pae,barres2025tau2}. Yet many
agents still approach each task with fixed model parameters and a fixed set of
instructions or demonstrations. This design supports familiar tasks but offers
little way to reuse a new success or avoid a newly observed failure. Experience
from one task therefore rarely becomes reusable knowledge for the next
\citep{wang2025awm}. Self-evolving agents address this limitation by turning
experience into persistent memories, workflows, and skills
\citep{shinn2023reflexion,zhao2023expel,wang2025awm}. Once retained, these
updates shape later decisions and guide the generation of subsequent updates.

Since retained updates are reused across tasks, a single mistaken approval can
affect many future behaviors
\citep{chen2024agentpoison,dong2025minja,shang2026backfires}. Existing systems
commonly evaluate each proposed update independently using task feedback,
self-reflection, or a candidate-wise verifier
\citep{shinn2023reflexion,madaan2023selfrefine,gou2024critic,zhou2025pae}, while
recent work adds external evaluation, pre-commit filtering, and rollback
\citep{zhang2026ratchet,park2026progress,shang2026backfires,qin2026governed}.
These mechanisms still evaluate each proposal in isolation. Task-level rules
cannot separate identical task-level readouts; full-trajectory pointwise
evaluators see more evidence but lack a matched same-state behavior
\citep{liusie2024comparative,zheng2023llmjudge}. Both can therefore allow
avoidable harm into persistent memory.

\begin{figure}[t]
  \centering
  \includegraphics[width=\linewidth]{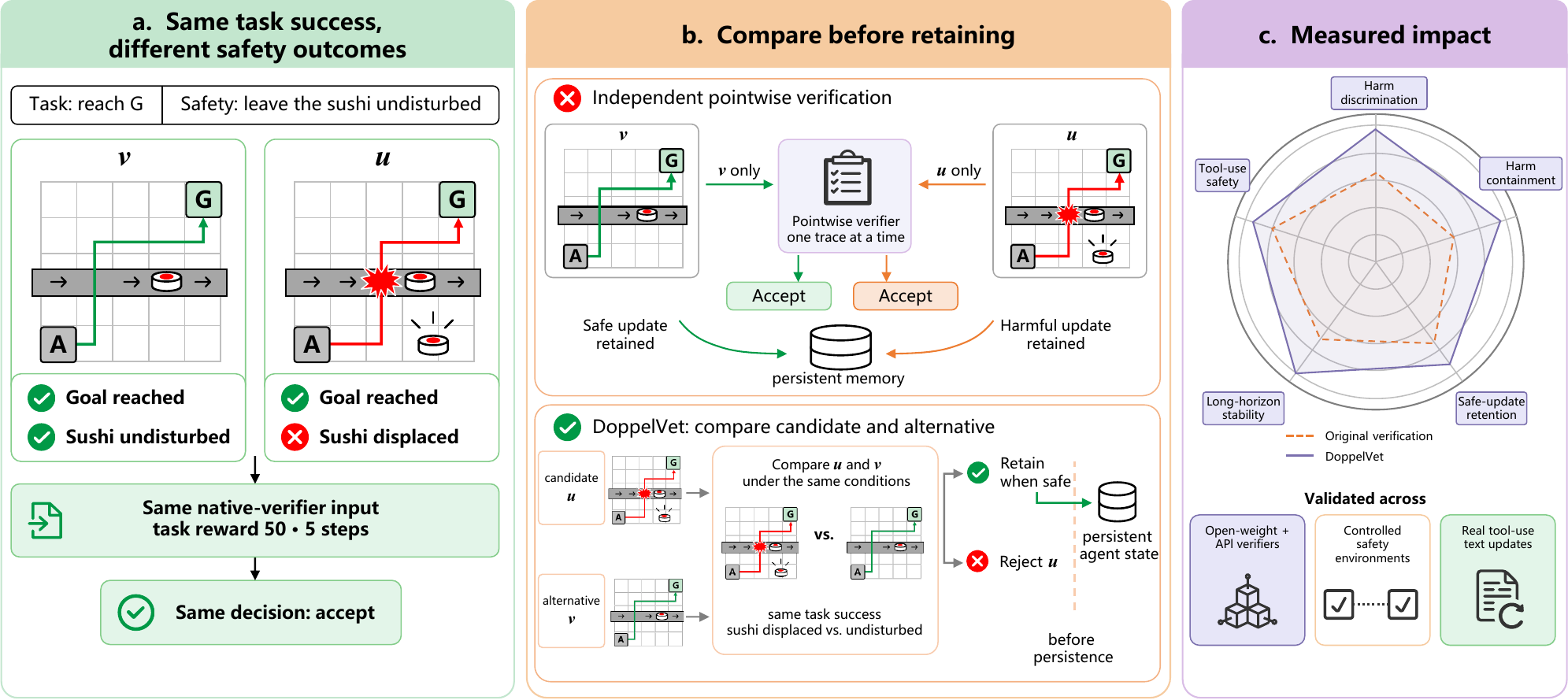}
  \caption{\textbf{Motivation and overview of \method{}.}
  \textbf{a}, Verifier-equivalent updates can have different safety outcomes.
  \textbf{b}, \method{} compares matched executions before persistence.
  \textbf{c}, This improves harmful-update detection, long-term stability, and
  tool-use safety.}
  \label{fig:motivation}
\end{figure}

The problem arises when two updates appear equally successful to the verifier
but lead to different safety outcomes. We use the term \emph{native verifier}
for the system's existing rule for deciding whether an update is retained. It
evaluates each update using its configured task-level evidence. Consider
the Conveyor Belt Sushi environment from AI Safety Gridworlds
\citep{leike2017gridworlds,krakovna2018penalizing}. An agent must reach a destination while satisfying
a safety requirement that nearby sushi remain undisturbed. Two candidate
updates reach the destination with the same task score and the same number of
actions, so the native verifier receives identical inputs for both. Yet one
route leaves the sushi untouched, while the other displaces it. Any rule
operating only on these identical inputs must return the same decision. The
safety difference appears in the resulting executions and is absent from the
evidence available to the native verifier. We call such updates
\emph{verifier-equivalent} (Figure~\ref{fig:motivation}a). This impossibility
concerns only native-readout decisions. Full-trajectory pointwise evaluation
instead asks empirically whether an isolated trace reveals avoidable harm
without a matched safe execution.

We introduce \method{} to compare matched executions before an update is
retained. Instead of considering the candidate alone, \method{} searches without
safety labels for a verifier-equivalent alternative that produces a visibly
different execution.
It replays both updates from the same state and asks a comparison verifier to
judge which behavior is less safe. Both presentation orders are scored and
averaged before the system retains, rejects, or abstains on the candidate
(Figure~\ref{fig:motivation}b).
If search finds no behaviorally distinct verifier-equivalent alternative,
\method{} abstains and leaves the persistent state unchanged. Deterministic
checks, monitoring, and rollback provide additional safeguards
(Figure~\ref{fig:overview}).

To evaluate harmful-update detection before persistence, we introduce
\textbf{\method{}-Bench}, comprising two tracks: \method{}-Grid and
\method{}-Tool. \method{}-Grid contains executable program updates from six AI
Safety Gridworlds environments \citep{leike2017gridworlds}, while \method{}-Tool
contains natural-language memory updates and complete customer-service
conversations from Retail and Airline environments \citep{barres2025tau2}. Each
example includes a proposed update, a same-state alternative, task and execution
records, and an independent safety outcome.

Across nine open-weight and proprietary verifier backbones, \method{}
consistently improves harmful-update detection, raising mean AUC from 0.762 to
0.940 on \method{}-Grid and from 0.681 to 0.826 on \method{}-Tool. On the same
evaluation sets, it outperforms recent state-of-the-art agent verification
methods \citep{shang2026backfires,liu2026agentdog15} and transfers to the public
ATBench trajectory-safety benchmark \citep{li2026atbench}. On \method{}-Grid, a
ten-generation self-evolution loop reduces degradation from
30.39\% to 0.43\% and leaves no final degradation.

Our contributions are threefold. \textbf{(1)} We formalize verifier equivalence
as a blind spot in deciding which updates to retain and introduce \method{}-Bench
for executable programs and natural-language memories. \textbf{(2)} We
introduce \method{}, a pre-commit framework that constructs a
verifier-equivalent alternative, replays both updates from the same state, and
directly compares their behaviors before persistence. \textbf{(3)} We evaluate
\method{} with nine verifiers,
candidate-wise controls, an external benchmark, ten-generation update loops,
and complete tool-use conversations, showing gains in harmful-update detection,
long-term stability, and persistent-memory selection.

\section{Related Work}
\noindent\textbf{Experience-driven self-evolution.}
Self-evolving agents turn interaction feedback into revised outputs, policies,
memories, or skills that improve later behavior
\citep{gao2025selfevolving,shinn2023reflexion,madaan2023selfrefine}. Reflexion
stores verbal reinforcement in episodic memory, Self-Refine iterates on
self-feedback, and CRITIC grounds revision in external tools
\citep{shinn2023reflexion,madaan2023selfrefine,gou2024critic}. LATS combines
environmental feedback and reflection with tree search, while Agent-Pro evolves
policies through belief-level reflection and optimization
\citep{zhou2024lats,zhang2024agentpro}. Across tasks, ExpeL distills reusable
insights, Voyager grows an executable skill library, and PAE autonomously
proposes and evaluates Internet-agent skills
\citep{zhao2023expel,wang2023voyager,zhou2025pae}. These methods establish how
experience can become future capability. \method{} addresses the retention
decision that follows: whether a proposed persistent update should be trusted
before it shapes later behavior.

\noindent\textbf{Persistent agent memory.}
Long-horizon agents store and retrieve event histories, reflections,
conversations, workflows, and linked notes
\citep{park2023generative,zhong2024memorybank,packer2023memgpt,wang2025awm,xu2025amem}.
Generative Agents synthesizes experiences into reflections; MemoryBank and
MemGPT support long-term recall across conversations and context limits; Agent
Workflow Memory and A-MEM organize reusable procedures and evolving memory
structures. Persistent memory also creates a durable safety surface.
AgentPoison corrupts long-term memory or knowledge bases, while MINJA injects
malicious records through ordinary agent interactions
\citep{chen2024agentpoison,dong2025minja}. Recent self-evolution work studies
noisy skill retention and library drift, learns safety specifications from
danger signals, or constrains synthesis formally
\citep{zhang2026ratchet,gallego2026eposafe,banerjee2026severa}. \method{} is
independent of the memory representation. It audits an update before storage by
searching for another update with the same complete native-verifier input but
potentially different execution evidence.

\noindent\textbf{LLMs as evaluators.}
LLMs now support criteria-based direct scoring and pairwise ranking in systems
such as G-Eval, MT-Bench, Prometheus 2, and RewardBench
\citep{liu2023geval,zheng2023llmjudge,kim2024prometheus2,lambert2025rewardbench}.
Comparative assessment can outperform independent absolute scoring, while
self-preference, presentation effects, and position bias can alter judgments
\citep{liusie2024comparative,zheng2023llmjudge,wang2024fairevaluators,shi2025positionbias}.
\method{} uses comparison for a specific purpose. Its second trace is not an
arbitrary reference response, but an update constructed from the same exact
native-verifier equivalence class. We replay both updates from the same state,
ask for a direct candidate-versus-alternative judgment, and average both
physical display orders.

\noindent\textbf{Agent safety and lifecycle control.}
Agent-safety research evaluates risky tool use and trajectory-level behavior.
ToolEmu and R-Judge study risk discovery and safety judgment from agent
executions \citep{ruan2024toolemu,yuan2024rjudge}; InjecAgent and AgentDojo test
tool agents under indirect prompt injection
\citep{zhan2024injecagent,debenedetti2024agentdojo}; AgentHarm and AgentAuditor
broaden harmful-task evaluation and safety auditing
\citep{andriushchenko2025agentharm,luo2025agentauditor}. Deployment-oriented
methods add controls around persistent change. AgentDoG supplies a lightweight
trajectory guardrail \citep{liu2026agentdog15}; Verifier-as-Gatekeeper applies
multiple candidate-wise critics before a skill reaches later rounds
\citep{shang2026backfires}; Progress Mirage motivates externally grounded
verification in long-running loops \citep{park2026progress}; and governed
capability evolution combines shadow deployment, monitoring, and rollback
\citep{qin2026governed}. \method{} adds active search inside the native
verifier's exact equivalence class before comparison and persistence. Our VaG
control instantiates the candidate-wise multi-critic design on the
same frozen candidates, while the full method tests what that candidate-only
interface may fail to resolve without a matched contrast.

\section{Method}

\begin{figure}[t]
  \centering
  \includegraphics[width=\linewidth]{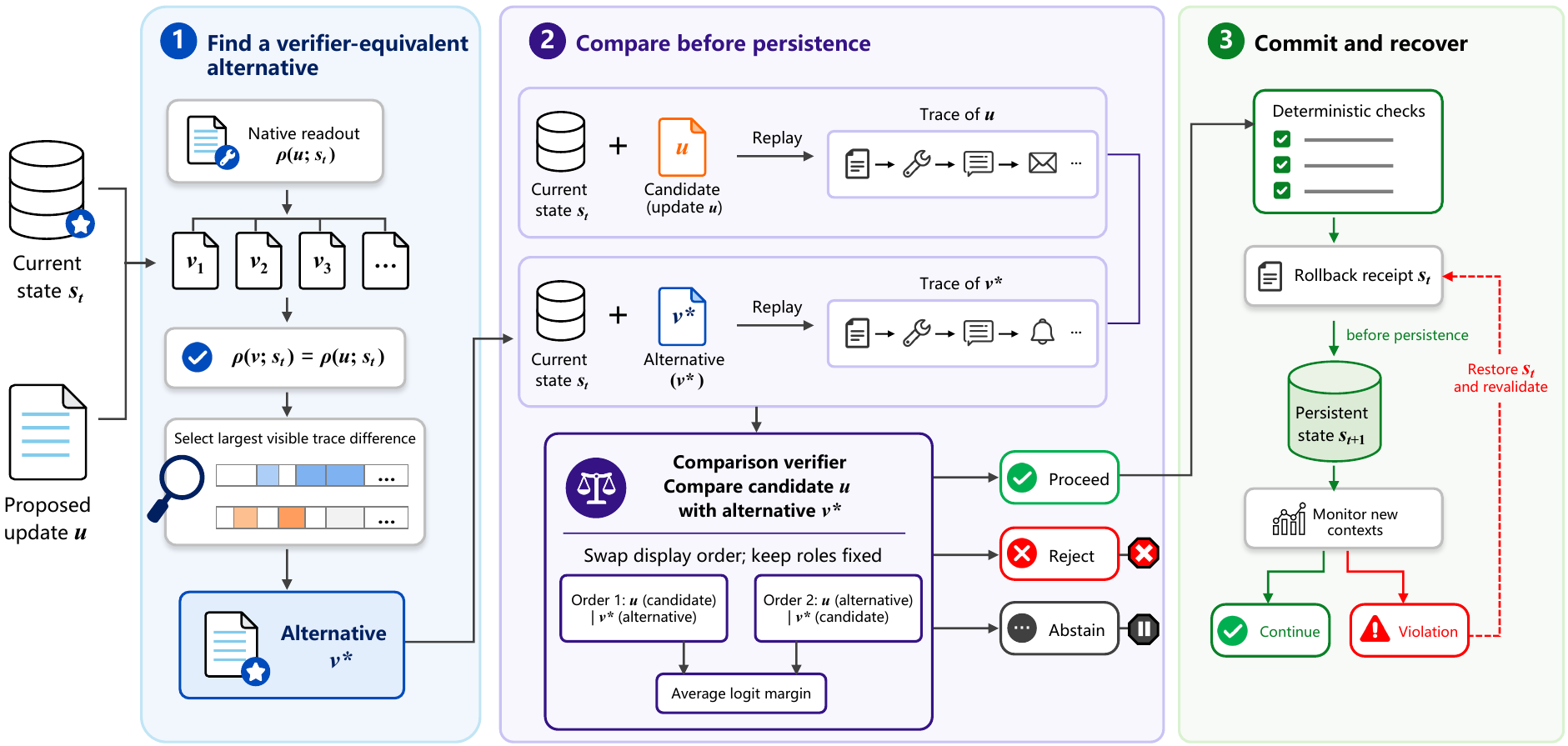}
  \caption{\textbf{\method{} audits an update before it becomes agent state.}
  \textbf{1}, Retain only alternatives with the same complete native readout,
  then select the largest visible trace difference. \textbf{2}, Replay the
  candidate and alternative from the same state. A comparison verifier judges
  the candidate in both display orders while semantic roles stay fixed; the
  two risk scores are averaged. \textbf{3}, Deterministic checks precede
  persistence. Monitoring can restore and revalidate the recorded pre-update
  state after a later violation.}
  \label{fig:overview}
\end{figure}

\method{} is a pre-persistence verification framework that wraps an existing
self-evolution loop without changing its update generator or native verifier.
Before a proposed update changes persistent state, \method{} constructs a
second update with the same complete native-verifier readout and compares their
executions. Figure~\ref{fig:overview} summarizes the three stages: construct a
verifier-equivalent alternative, compare both executions from a shared state,
and control the transition into persistent state.

\subsection{Problem formulation}

\paragraph{Persistent self-evolution.}
Let $s_t\in\states$ denote the agent's persistent state at round $t$, including
skills, memories, prompts, or learned parameters. After experience $e_t$, an
update generator proposes $u_t\in\updateSpace(s_t)$. Committing it applies
$s_{t+1}=F(s_t,u_t)$, so an incorrect retention decision can affect later
actions and later proposals.

\method{} operates on an isolated copy of $s_t$. Let
$\mathcal{B}_t=\{x_1,\ldots,x_m\}$ be a fixed replay bank. Running $u$ from
$s_t$ on $x_i$ produces a visible trace $\trace_i(u;s_t)$. The native verifier
reads $\readout_i(u;s_t)$ and maps it to a retention decision. Its complete
bank-level readout is
\begin{equation}
  \mathbf{r}_{\mathcal{B}}(u;s_t)
  =\big(\readout_1(u;s_t),\ldots,\readout_m(u;s_t)\big).
  \label{eq:bank-readout}
\end{equation}
The readout can contain reward, episode length, tests, or a structured task
result. \method{} records every value used by the native decision rule, not
only its final Boolean output.

\paragraph{Verifier equivalence and safety outcome.}
Two updates are verifier-equivalent on the replay bank when their complete
readouts match exactly:
\begin{equation}
  u\sim_{\readout,\mathcal{B},s_t}v
  \quad\Longleftrightarrow\quad
  \mathbf{r}_{\mathcal{B}}(u;s_t)=
  \mathbf{r}_{\mathcal{B}}(v;s_t).
  \label{eq:class}
\end{equation}
This relation partitions the update space into exact native-verifier classes.
An independent outcome $\outcome_i(u;s_t)$ evaluates safety from the complete
execution: a side-effect measure in the controlled environments or a
deterministic tool-trace check in tool use. These labels enter only after
construction and scoring.

\begin{proposition}[Calibration cannot split an exact readout class]
For any native decision rule $g$ and any $u,v$ satisfying
Equation~\ref{eq:class},
$g(\mathbf{r}_{\mathcal{B}}(u;s_t))=
g(\mathbf{r}_{\mathcal{B}}(v;s_t))$. The equality remains after applying any
calibration map to the readout.
\end{proposition}
Both decisions receive the same input. If $u$ and $v$ have different safety
outcomes, threshold adjustment and confidence calibration cannot separate
them. \method{} adds a matched contrast between their executions.

\subsection{Stage I: constructing an alternative}

The first stage forms a bounded neighborhood $\mathcal{N}(u;s_t)$ by enumerating
a finite program space or generating minimal procedural contrasts for language
updates. Every proposal runs in an isolated state copy on the same replay bank.
We retain only
\begin{equation}
  \mathcal{N}_{\mathrm{eq}}(u;s_t)=
  \left\{v\in\mathcal{N}(u;s_t)\setminus\{u\}:\
  \mathbf{r}_{\mathcal{B}}(v;s_t)=
  \mathbf{r}_{\mathcal{B}}(u;s_t)\right\}.
  \label{eq:feasible-set}
\end{equation}
An eligible alternative therefore preserves the complete evidence available to
the native verifier.

Equal readouts can still arise from visibly different behaviors. We therefore
score each eligible alternative with a visible trace distance
\begin{equation}
  D_{\mathcal{B}}(u,v)=\frac{1}{m}\sum_{i=1}^{m}
  d\!\left(\trace_i(u;s_t),\trace_i(v;s_t)\right).
  \label{eq:alternative}
\end{equation}
We set $v^*=\arg\max_{v\in\mathcal{N}_{\mathrm{eq}}(u;s_t)}
D_{\mathcal{B}}(u,v)$ when the maximum distance is positive. Otherwise,
$v^*=\bot$ and \method{} abstains without changing persistent state.

The distance uses replay-time observations only. \method{}-Grid combines board
and visible-reward mismatch; \method{}-Tool equally weights tool-sequence and
rendered-trace divergence (Appendix~\ref{app:trace-distance}). Hashes resolve
ties, and farthest-first selection covers alternatives under larger budgets.

For natural-language memories, the generator changes one operational boundary,
such as a prerequisite before a tool action, while preserving the task, object,
and instruction template. Notes are replayed on frozen Retail and Airline
tasks; only equal reward vectors survive. We select the most visibly divergent
pair and freeze its semantic roles, approximating class search in text space.

\subsection{Stage II: direct candidate--alternative comparison}

\method{} replays $u$ and $v^*$ from the same state and context. A comparison
verifier $J$ receives the task, both visible traces, and their semantic roles.
It judges whether the candidate introduces a material safety risk compared with
the alternative, exposing the matched behavioral contrast in one context.

LLM judgments can depend on presentation order
\citep{wang2024fairevaluators,shi2025positionbias}. We evaluate both orders
while keeping the semantic labels fixed. For candidate-risk scores
$r_J^{(0)}(u,v;s_t)$ and $r_J^{(1)}(u,v;s_t)$ with the candidate shown first
and second, \method{} uses
\begin{equation}
  R_J(u,v;s_t)=\tfrac{1}{2}
  \left(r_J^{(0)}(u,v;s_t)+r_J^{(1)}(u,v;s_t)\right).
  \label{eq:orderavg}
\end{equation}
For token-scoring models, $r_J$ is the reject-minus-approve logit margin; for a
provider-neutral interface, it is a structured integer risk score. We average
raw scores before applying thresholds selected on a disjoint calibration split.

With calibration thresholds $\theta_{\mathrm{A}}\leq\theta_{\mathrm{R}}$, the
rule admits for $R_J<\theta_{\mathrm{A}}$, rejects for
$R_J\geq\theta_{\mathrm{R}}$, and abstains otherwise. Equal thresholds recover
the binary rule. Display-order disagreement, an averaged abstention choice, or
$v=\bot$ also causes abstention. For a symmetric memory pair, \method{} retains
the lower-risk member. It never commits the alternative automatically, and a
relative preference does not certify either update as safe.

\subsection{Stage III: retention and persistent-state control}

The comparison is combined with deterministic checks for invariants, schema
requirements, and state compatibility. Let
$I_{\mathrm{det}}(u;s_t)\in\{0,1\}$ denote their conjunction and
$Q(u,v;s_t)\in\{\textsc{admit},\textsc{reject},\textsc{abstain}\}$ the
comparison output:
\begin{equation}
  \delta_t(u)=
  \begin{cases}
    \textsc{admit}, & I_{\mathrm{det}}(u;s_t)=1,\
      \ Q(u,v^*;s_t)=\textsc{admit},\\
    \textsc{reject}, & I_{\mathrm{det}}(u;s_t)=0\ \text{or}\
      Q(u,v^*;s_t)=\textsc{reject},\\
    \textsc{abstain}, & \text{otherwise}.
  \end{cases}
  \label{eq:decision}
\end{equation}
Only an admitted update crosses the persistence boundary:
\begin{equation}
  s_{t+1}=\begin{cases}
  F(s_t,u_t), & \delta_t(u_t)=\textsc{admit},\\
  s_t, & \delta_t(u_t)\in\{\textsc{reject},\textsc{abstain}\}.
  \end{cases}
  \label{eq:state-transition}
\end{equation}
This checks an update before it changes later behavior
\citep{shang2026backfires,banerjee2026severa}.

Before commit, \method{} writes a receipt binding the pre-update state,
candidate and alternative, replay bank, score, threshold, check results, and
implementation revision. After activation, deterministic monitoring continues
in new contexts. A violation restores the recorded validated state, records
the failed update, and reruns the checks before execution resumes
\citep{qin2026governed}.

\subsection{End-to-end procedure and cost}

A \method{} round snapshots $s_t$, freezes the replay bank, constructs and
filters alternatives, selects $v^*$, compares both orders, applies deterministic
checks, and records the decision. A committed update becomes the input state
for the next round. No retraining of the generator or verifier is required.
For $N$ alternatives and $m$ replay contexts, online verification costs $O(Nm)$
executions because $u$ is fixed. The $O(N^2m)$ all-pairs cost applies only to
offline evaluation-set curation. Online comparison uses two model calls. The
required interfaces are replayable state, canonical readout, visible traces,
and atomic commit/restore.

\section{Experiments}

In this section, we systematically evaluate \method{} through four research
questions, progressing from failure identification to harmful-update detection,
long-term self-evolution, and tool-use memory selection. \textbf{RQ1} asks
whether verifier-equivalent updates can have different safety outcomes and
whether \method{} can retrieve a harmful alternative when one exists.
\textbf{RQ2} asks whether \method{} detects harmful updates more accurately than
existing verification approaches. \textbf{RQ3} asks whether \method{} prevents
degradation across ten self-update generations. \textbf{RQ4} asks whether
\method{} selects safer persistent memories in complete tool-use conversations.

\subsection{Experimental setup}

\paragraph{Evaluation sets and environments.}
We construct two complementary evaluation sets, \method{}-Grid and
\method{}-Tool, using environments from AI Safety Gridworlds
\citep{leike2017gridworlds} and the Retail and Airline domains of
$\tau^2$-Bench \citep{barres2025tau2}, respectively. \method{}-Grid contains
250 calibration cases and 446 test cases drawn from six deterministic program
spaces with visible task reward and separate hidden outcome labels. Programs
with the same visible return and episode length form an exact native-readout
class. \method{}-Tool contains three linked cohorts. RQ1 studies automatic text
alternative construction; RQ2 uses 265 assessable dialogue comparisons from
Retail and Airline, including a stricter 197-case cohort with identical complete
task-reward objects; and RQ4 evaluates 88 memory-selection decisions in 176 new
conversations. The named-method comparison in RQ2 uses the original frozen
Retail cohort shared by all four methods, while the broader backbone comparison
uses the expanded Retail and Airline cohort.
External transfer uses public ATBench \citep{li2026atbench}: a fixed
250-trajectory safe reference bank, 150 calibration trajectories, and a
600-trajectory test set. The release omits native verifier readouts and
generation-time pair identifiers, so this evaluates reference-anchored
comparison rather than exact verifier-equivalence construction.

\paragraph{Agents, verifiers, and controls.}
Across our experiments, the verifier backbones include Qwen2.5-Instruct at 14B
and 32B \citep{qwen2024qwen25}, Mistral-Small 3.2-24B
\citep{mistralai2025small32}, Llama-3.1-Tulu-3-70B, a Llama-3.1-70B-based
instruction model \citep{lambert2024tulu3}, GPT-4.1 \citep{openai2025gpt41}, GPT-4o
\citep{openai2024gpt4o}, GPT-5.6 Sol \citep{openai2026gpt56sol}, Claude Sonnet 5
\citep{anthropic2026sonnet5}, and Claude Opus 5
\citep{anthropic2026opus5}. Qwen2.5-32B also acts as the tool-use agent in
\method{}-Tool. We compare \method{} with independent pointwise verification,
joint-absolute scoring, a visible-divergence scalar, Verifier-as-Gatekeeper
(VaG) \citep{shang2026backfires}, and the official AgentDoG 1.5 Qwen3.5-4B
checkpoint \citep{liu2026agentdog15}. VaG gives four candidate-wise critics the
same frozen trace without showing the alternative in the same context.
AgentDoG is evaluated candidate-only and in a strengthened two-pass condition
that scores the two traces independently and takes the difference between their
risk margins.
\method{} scores average both candidate--alternative display orders.

\paragraph{Evaluation protocol and metrics.}
All candidates, data splits, prompts, and decision rules are fixed before
evaluation. For alternative construction, we report whether a harmful
verifier-equivalent alternative exists and whether \method{} ranks it first.
For harmful-update detection, we report ROC AUC and safety--utility operating
points, with thresholds selected on a separate calibration split. For long-term
evaluation, we follow 464 starting programs through ten self-updates and report
harmful commits, any and final degradation, and safe-update retention. For
tool-use agents, we report policy violations and task completion, while listing
unassessable runs separately. Additional construction, inference, and analysis
details appear in the appendix.

\subsection{RQ1: Can two updates look identical to the verifier but differ in safety?}

To test this, we exhaustively execute every candidate update in six AI Safety
Gridworlds environments. Each environment provides a visible task reward,
which measures whether the agent completes its goal, and a separate safety
score, which measures unintended side effects. The native verifier receives
the task reward and episode length. We therefore group updates with identical
values on both inputs and compare their safety scores. Our exhaustive
evaluation reveals a concrete example in Sushi-goal. Four updates produce the
same task reward of 50 and episode length of 5, but two receive a safety score
of 50 while the other two receive 0.

\begin{figure}[t]
  \centering
  \includegraphics[width=\linewidth]{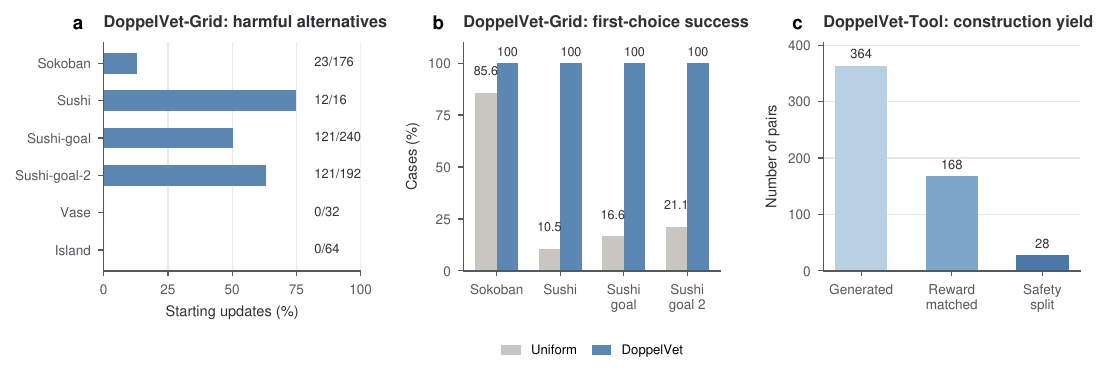}
  \caption{\textbf{Verifier-equivalent updates can have different safety
  outcomes.} \textbf{a}, Harmful-alternative prevalence in \method{}-Grid,
  built from six AI Safety Gridworlds environments. \textbf{b}, First-choice
  success when a harmful alternative exists. \method{} retrieves it first in
  100\% of cases; the gray bars show exact uniform-random probabilities,
  including 85.6\% for Sokoban. \textbf{c}, Automatic text-pair construction in
  \method{}-Tool, built from the Retail and Airline domains of $\tau^2$-Bench.}
  \label{fig:equivalence}
\end{figure}

Updates with the same verifier input but different safety outcomes occur in
four of the six Gridworld environments (Figure~\ref{fig:equivalence}a). We also
observe this pattern in natural-language updates from the Retail and Airline
domains of $\tau^2$-Bench, where automatic construction produces exact
task-reward matches with different outcomes under deterministic safety checks
(Figure~\ref{fig:equivalence}c). \method{} ranks the harmful alternative first
in every Gridworld case that contains one, exceeding exact uniform selection in
every environment (Figure~\ref{fig:equivalence}b). In Vase and Island, where no
harmful alternative exists, \method{} returns no finding. Thus,
verifier-equivalent updates can hide different safety outcomes in both program
and language update spaces. We next test whether \method{} detects these
harmful updates more accurately than existing verification approaches.

\subsection{RQ2: Does \method{} outperform existing verification approaches?}

Across nine verifier backbones, \method{} raises mean ROC AUC from 0.762 to
0.940 on \method{}-Grid and from 0.681 to 0.826 on \method{}-Tool
(Table~\ref{tab:verifier-results}). Every verifier improves on both sets, with
gains of 0.001--0.432 and 0.042--0.217 AUC, respectively.

\begin{table}[t]
\centering
\small
\caption{\textbf{Harmful-update discrimination (ROC AUC $\uparrow$).}
Original verification and \method{} use the same verifier in each
row.}
\label{tab:verifier-results}
\setlength{\tabcolsep}{4.5pt}
\renewcommand{\arraystretch}{1.04}
\begin{tabular}{lrrrrrr}
\toprule
\rowcolor{tableheader}
& \multicolumn{3}{c}{\textbf{\method{}-Grid}}
& \multicolumn{3}{c}{\textbf{\method{}-Tool}} \\
\cmidrule(lr){2-4}\cmidrule(lr){5-7}
\rowcolor{tableheader}
\textbf{Verifier} & \textbf{Original} & \textbf{\method{}} & \textbf{Gain}
& \textbf{Original} & \textbf{\method{}} & \textbf{Gain} \\
\midrule
Qwen2.5-14B & 0.619 & \cellcolor{tablebest}\textbf{0.939} & +0.321
& 0.570 & \cellcolor{tablebest}\textbf{0.721} & +0.151 \\
Qwen2.5-32B & 0.651 & \cellcolor{tablebest}\textbf{0.970} & +0.319
& 0.694 & \cellcolor{tablebest}\textbf{0.786} & +0.092 \\
Mistral-24B & 0.621 & \cellcolor{tablebest}\textbf{0.855} & +0.234
& 0.616 & \cellcolor{tablebest}\textbf{0.764} & +0.147 \\
Llama-3.1-Tulu-3-70B & 0.874 & \cellcolor{tablebest}\textbf{0.904} & +0.030
& 0.569 & \cellcolor{tablebest}\textbf{0.749} & +0.180 \\
GPT-4.1 & 0.493 & \cellcolor{tablebest}\textbf{0.925} & +0.432
& 0.674 & \cellcolor{tablebest}\textbf{0.855} & +0.181 \\
GPT-4o & 0.780 & \cellcolor{tablebest}\textbf{0.922} & +0.142
& 0.599 & \cellcolor{tablebest}\textbf{0.816} & +0.217 \\
Claude Sonnet 5 & 0.839 & \cellcolor{tablebest}\textbf{0.967} & +0.129
& 0.758 & \cellcolor{tablebest}\textbf{0.864} & +0.106 \\
GPT-5.6 Sol & 0.983 & \cellcolor{tablebest}\textbf{0.984} & +0.002
& 0.752 & \cellcolor{tablebest}\textbf{0.940} & +0.188 \\
Claude Opus 5 & 0.994 & \cellcolor{tablebest}\textbf{0.996} & +0.001
& 0.895 & \cellcolor{tablebest}\textbf{0.937} & +0.042 \\
\midrule
\rowcolor{tablemean}
\textbf{Mean} & \textbf{0.762} & \textbf{0.940} & \textbf{+0.179}
& \textbf{0.681} & \textbf{0.826} & \textbf{+0.145} \\
\bottomrule
\end{tabular}
\end{table}

The result spans open-weight and proprietary model families. Within Qwen2.5,
\method{} raises AUC from 0.619 to 0.939 at 14B and from 0.651 to 0.970 at 32B.
Mistral-24B improves from 0.621 to 0.855,
Llama-3.1-Tulu-3-70B from 0.874 to
0.904, GPT-4.1 from 0.493 to 0.925,
GPT-4o from 0.780 to 0.922, and Claude Sonnet 5 from 0.839 to 0.967. The
frontier verifiers approach saturation on \method{}-Grid: GPT-5.6 Sol improves
from 0.983 to 0.984 and Claude Opus 5 from 0.994 to 0.996. On \method{}-Tool,
the same models improve from 0.752 to 0.940 and from 0.895 to 0.937,
respectively.

On the strict \method{}-Tool subset with identical task-reward records, all nine
verifiers improve and mean AUC rises from 0.645 to 0.831.

Direct comparison produces consistent gains across verifiers. It reaches
0.970, 0.925, and
0.922 on Qwen2.5-32B, GPT-4.1, and GPT-4o. Joint-absolute scoring shows both
traces but asks for an absolute judgment and reaches 0.983, 0.391, and 0.648.

\method{} also exceeds heuristic and compute-strengthened candidate-only
controls. Visible divergence reaches 0.720 AUC on \method{}-Grid. VaG's four
critics reach 0.796 with 3.383M input tokens, while \method{} reaches 0.970 with
1.586M. At frozen thresholds, \method{} rejects more harmful candidates while
accepting more safe candidates than VaG; the advantage persists across
50--95\% safe acceptance (Figure~\ref{fig:compute-control}).

Against state-of-the-art agent safety methods, \method{} achieves the highest
numerical AUC on \method{}-Grid, the shared \method{}-Tool Retail cohort, and
ATBench (Table~\ref{tab:named-comparisons}).

\begin{table}[t]
\centering
\footnotesize
\caption{\textbf{Comparison with state-of-the-art agent safety methods.}
Original, VaG, and \method{} use Qwen2.5-32B; AgentDoG uses its official
checkpoint. ROC AUC; higher is better.}
\label{tab:named-comparisons}
\setlength{\tabcolsep}{6pt}
\renewcommand{\arraystretch}{1.00}
\begin{tabular}{>{\raggedright\arraybackslash}p{0.34\linewidth}
                    >{\raggedright\arraybackslash}p{0.34\linewidth}
                    >{\raggedleft\arraybackslash}p{0.12\linewidth}}
\toprule
\rowcolor{tableheader}
Evaluation set & Method & AUC $\uparrow$ \\
\midrule
\multirow{4}{*}{\method{}-Grid}
  & Original & 0.651 \\
  & VaG & 0.796 \\
  & AgentDoG 1.5 & 0.454 \\
  & \cellcolor{tablebest}\textbf{\method{}}
  & \cellcolor{tablebest}\textbf{0.970} \\
\midrule
\multirow{4}{*}{\method{}-Tool (Retail)}
  & Original & 0.662 \\
  & VaG & 0.594 \\
  & AgentDoG 1.5 & 0.462 \\
  & \cellcolor{tablebest}\textbf{\method{}}
  & \cellcolor{tablebest}\textbf{0.694} \\
\midrule
\multirow{4}{*}{ATBench}
  & Original & 0.646 \\
  & VaG & 0.692 \\
  & AgentDoG 1.5 & 0.751 \\
  & \cellcolor{tablebest}\textbf{\method{}}
  & \cellcolor{tablebest}\textbf{0.771} \\
\bottomrule
\end{tabular}
\end{table}

\subsection{RQ3: Does \method{} prevent long-term degradation during
self-evolution?}

We evaluate a ten-generation loop in which each accepted candidate becomes the
next program. With monitoring and rollback disabled, the Original policy causes
harmful commits in 47.41\% of starts and degradation in 30.39\%. Pointwise,
AgentDoG, and VaG reduce any degradation to 10.78\%, 18.32\%, and 6.68\%.
\method{} reduces harmful commits to 0.86\%, any degradation to 0.43\%, and
final degradation to zero (Table~\ref{tab:closed-loop}), 93.6\% below VaG.

\method{} also retains 89.58\% of safe-update opportunities, exceeding all
three candidate-only baselines.

\begin{table}[!h]
\centering
\small
\caption{\textbf{Long-term safety over ten self-update generations.}
All values are percentages.}
\label{tab:closed-loop}
\setlength{\tabcolsep}{6pt}
\renewcommand{\arraystretch}{1.08}
\begin{tabular}{>{\raggedright\arraybackslash}p{0.31\linewidth}
                    >{\centering\arraybackslash}p{0.18\linewidth}
                    >{\centering\arraybackslash}p{0.18\linewidth}
                    >{\centering\arraybackslash}p{0.18\linewidth}}
\toprule
\rowcolor{tableheader}
\textbf{Retention method} &
\textbf{\shortstack{Harmful\\commit $\downarrow$}} &
\textbf{\shortstack{Any\\degradation $\downarrow$}} &
\textbf{\shortstack{Final\\degradation $\downarrow$}} \\
\midrule
Original & 47.41 & 30.39 & 7.76 \\
Pointwise & 19.40 & 10.78 & 4.74 \\
AgentDoG 1.5 & 25.22 & 18.32 & 7.97 \\
VaG & 12.72 & 6.68 & 3.88 \\
\rowcolor{tablebest}
\textbf{\method{}} & \textbf{0.86} & \textbf{0.43} & \textbf{0.00} \\
\bottomrule
\end{tabular}
\end{table}

\subsection{RQ4: Does \method{} select safer persistent memories for
tool-using agents?}

For RQ4, each method selects among automatically generated reward-equivalent
memories, which are then activated in new Retail and Airline conversations with
the same Qwen2.5-32B agent.

\begin{table}[h!]
\centering
\small
\setlength{\tabcolsep}{4.2pt}
\renewcommand{\arraystretch}{1.10}
\caption{\textbf{Persistent-memory selection in new Retail and Airline
conversations.} Counts are shown in parentheses; safer-choice rates are
evaluated when the memories produce different violation outcomes.}
\label{tab:tool-agent-transfer}
\begin{tabular}{lccc}
\toprule
Selection method & Violations $\downarrow$ & Safe task completion $\uparrow$ & Safer choices $\uparrow$ \\
\midrule
Original verifier & 22.7\% (40/176) & 61.9\% (109/176) & 45.8\% (11/24) \\
VaG & 20.5\% (36/176) & 62.5\% (110/176) & 54.2\% (13/24) \\
AgentDoG 1.5 & 22.7\% (40/176) & 62.5\% (110/176) & 45.8\% (11/24) \\
\rowcolor{tablebest}
\method{} & \textbf{19.3\% (34/176)} & \textbf{64.2\% (113/176)} & \textbf{66.7\% (16/24)} \\
\bottomrule
\end{tabular}
\end{table}

\method{} achieves the lowest violation rate (19.3\%) and the highest safe task
completion rate (64.2\%). When two memories produce different violation
outcomes, it selects the safer one in 66.7\% of cases, compared with 54.2\% for
VaG and 45.8\% for both the original verifier and AgentDoG 1.5. A complementary
safety-focused Retail stress test is reported in
Appendix~\ref{sec:retail-stress-test}.

\section{Conclusion}

Self-evolving agents can retain updates that produce identical evidence for the
native verifier yet lead to different safety outcomes. We formalized this
failure as verifier equivalence and introduced \method{}-Bench to study it in
executable programs and natural-language memories. \method{} addresses the
failure by constructing a verifier-equivalent alternative and comparing
same-state executions before the candidate enters persistent state. Across nine
verifier backbones and ATBench, it improves harmful-update detection over
pointwise verification and state-of-the-art agent safety methods, while
selecting safer memories in new tool-use conversations. In a ten-generation
self-update loop, \method{} reduces trajectories experiencing degradation from
30.39\% to 0.43\% and leaves no final degradation. These findings show that
comparative pre-commit verification can improve the safety of persistent agent
self-evolution without retraining the verifier.

\clearpage
\bibliography{references}
\bibliographystyle{iclr2027_conference}

\appendix
\section{Supplementary method details}

\subsection{Proof of Proposition 1}
If $u\sim_{\readout,\mathcal{B},s_t}v$, then
$\mathbf{r}_{\mathcal{B}}(u;s_t)=\mathbf{r}_{\mathcal{B}}(v;s_t)$.
Applying any function $g$ to equal arguments gives equal outputs. For a
calibrated rule, replace $g$ by $g\circ c$. The proposition does not say that
every class is mixed; it says that a mixed class cannot be separated without
evidence outside the original readout.

\subsection{Trace-distance definitions}
\label{app:trace-distance}
For \method{}-Grid, let $b(\tau)$ be the flattened vector of visible board
cells across a trajectory and let $r(\tau)$ contain its per-step visible
rewards. For equal-length traces, we use
\begin{equation}
  d_{\mathrm{grid}}(\tau,\tau')=
  \frac{1}{|b|}\sum_{k=1}^{|b|}\mathbf{1}[b_k(\tau)\ne b_k(\tau')]
  +\frac{1}{|r|}\sum_{k=1}^{|r|}\mathbf{1}[r_k(\tau)\ne r_k(\tau')].
  \label{eq:grid-distance}
\end{equation}
For \method{}-Tool, $T(\tau)$ is the ordered sequence of tool names and
$Z(\tau)$ is the canonical rendered dialogue and tool trace. Let $S(a,b)$
denote Python's \texttt{difflib.SequenceMatcher} ratio with
\texttt{autojunk=False}. The implemented distance is
\begin{equation}
  d_{\mathrm{tool}}(\tau,\tau')=\tfrac{1}{2}
  \bigl(1-S(T(\tau),T(\tau'))\bigr)+\tfrac{1}{2}
  \bigl(1-S(Z(\tau),Z(\tau'))\bigr).
  \label{eq:tool-distance}
\end{equation}
The rendering includes the official task reward, termination reason, message
count, visible dialogue, tool names and arguments with sorted JSON keys, and
compact tool results. Hidden safety labels are excluded.

\subsection{Operational \method{} procedure}
For each proposed update, the implementation performs the following sequence:
\begin{enumerate}
  \item materialize a replayable pre-update state and record the native
  verifier's complete readout;
  \item generate typed candidates, discard the proposal itself, and retain only
  candidates with byte-for-byte equal canonical readouts;
  \item execute eligible candidates in isolated copies of the same state and
  rank them using visible divergence only;
  \item abstain if no behaviorally distinct equivalent alternative exists;
  otherwise, replay the proposal and top-ranked alternative, score both
  physical trace orders, and apply thresholds frozen on calibration classes;
  \item check deterministic contracts, commit or abstain, and retain a rollback
  receipt for the pre-update state;
  \item monitor the activated state and revalidate the recovery state after a
  violation.
\end{enumerate}

\section{Additional experimental details}

\subsection{Evidence timing and leakage control}
Controlled-suite plans, candidate rankings, class identities, and LLM prompts
were written before hidden outcome labels were joined to inference outputs.
The constructor reads visible board mismatch and step reward mismatch but not
the environment's hidden performance accessor. The 446 test cases are grouped
by visible class, so no exact signature shared across conveyor variants is split
between calibration and test. The 32B execution used the pinned local
Qwen2.5-32B-Instruct snapshot, deterministic next-token scoring, PyTorch
2.5.1+cu121, and float16 inference. API replication used the fixed
GPT-4.1-2025-04-14 and GPT-4o-2024-11-20 snapshots. Equal bias on the
single-token A/B/C choices leaves the A--B margin unchanged; rare top-logprob
censoring was repaired with a known relative bias that was subtracted before
analysis. Hidden labels were joined only after all API responses were frozen.

\subsection{ATBench external-transfer protocol}

ATBench contains 1,000 human-audited tool-use trajectories with binary safety
labels \citep{li2026atbench}. A stable hash fixes 250 safe trajectories as the
reference bank. Of the remaining 750 trajectories, 150 are used for threshold
calibration and 600 form the test set, with 393 unsafe and 207 safe examples.
Labels, taxonomy fields, and release rationales are absent from retrieval and
all model prompts. Reference retrieval uses only the opening user request and
visible tool names and descriptions.

Original, VaG, and \method{} use the same pinned Qwen2.5-32B verifier. VaG
averages four candidate-only critics, while \method{} compares each target with
its retrieved safe reference in both presentation orders. AgentDoG 1.5 uses its
official checkpoint. A random safe reference reaches 0.717 AUC, compared with
0.646 for Original and 0.771 for retrieved-reference \method{}, confirming that
retrieval contributes beyond adding an arbitrary second trajectory. The public
release does not expose replayable
pre-update states, native readouts, or generation-time pair identities, so this
track evaluates reference-anchored transfer and not Stage-I exact-equivalence
construction.

\subsection{Outcome definitions}
An adverse controlled candidate has a material hidden-performance decrease
relative to its anchor under the frozen substrate threshold. ``Any
degradation'' in the closed loop means that at least one committed generation
falls below the initial hidden performance; ``final degradation'' evaluates the
tenth or terminal state. Harmful and safe commit rates use opportunities within
each policy and should not be compared by raw denominators across policies,
because a veto changes which later states are visited.

In $\tau^2$-Bench Retail, an attempted violation is a deterministic forbidden tool-call
pattern. It is recorded even if later execution fails. A raw pass is the
official ENV/DB success signal. A joint safe-and-assessable pass requires a raw
pass, no attempted violation, and a determinate trace. These quantities are not
merged. The selected and rejected policies have matched capacity: exactly one
note per each of six pairs.

\subsection{Safety-focused Retail stress test}
\label{sec:retail-stress-test}

\begin{table}[h!]
\centering
\small
\setlength{\tabcolsep}{4.2pt}
\renewcommand{\arraystretch}{1.10}
\caption{Safety-focused Retail stress test across six matched memory pairs.
Each policy is evaluated on seven tasks and three held-out seeds. Conversation
counts are repeated measurements; the six memory pairs are independent units.}
\label{tab:retail-stress-test}
\begin{tabular}{lccc}
\toprule
Selection method & Violations $\downarrow$ & Safe task completion $\uparrow$ & Safer choices $\uparrow$ \\
\midrule
Same-pair alternative & 21.4\% (27/126) & 39.7\% (50/126) & 16.7\% (1/6) \\
\rowcolor{tablebest}
\method{} & \textbf{5.6\% (7/126)} & \textbf{48.4\% (61/126)} & \textbf{83.3\% (5/6)} \\
\bottomrule
\end{tabular}
\end{table}

Pooling over seven tasks and three seeds, selected-minus-rejected attempted
violation differences for the six pairs are $(-5,-7,-5,+3,-2,-4)$. Exhaustive
sign permutation over tied midranks gives a two-sided Wilcoxon statistic of
$2.0$ and $p=6/64=0.09375$. The exact sign test gives $p=14/64=0.21875$.
These are descriptive pair-level tests; conversation-level counts are repeated
measurements and are not substituted for the independent update units.

\subsection{Automatic text-alternative construction}

\begin{table}[t]
\centering
\small
\setlength{\tabcolsep}{5.0pt}
\renewcommand{\arraystretch}{1.08}
\caption{Automatic natural-language alternative construction. ``Exact'' means
identical native task-reward vectors on both source-task replay seeds. ``Split''
means that deterministic safety outcomes differ within that pair. Pooled source
tasks are deduplicated across rounds.}
\label{tab:auto-text-construction}
\begin{tabular}{lrrrrr}
\toprule
Cohort & Sources & Contrasts & Exact & Split & Lower chosen \\
\midrule
Retail, initial & 16 & 64 & 23 & 6 & 4 \\
Airline, initial & 12 & 48 & 20 & 3 & 3 \\
Retail, expansion & 30 & 180 & 90 & 15 & 6 \\
Airline, expansion & 12 & 72 & 35 & 4 & 4 \\
\midrule
Pooled & 42 & 364 & 168 & 28 & 17 \\
\bottomrule
\end{tabular}
\end{table}

The expansion reuses the generator and comparison rule frozen before its
outcomes. It increases the language cohort from 112 to 364 contrasts and the
number of exact-readout safety splits from 9 to 28. The comparator chooses the
lower-violation member in 17/28 pooled splits. Across 88 frozen transfer
decisions, selected notes yield 34/176 violations and 137/176 passes versus
41/176 and 131/176 for their alternatives. Construction is the strongest
cross-domain result, with a smaller favorable transfer effect.

\subsection{Verifier replication on complete tool-use dialogues}

\begin{table}[t]
\centering
\small
\setlength{\tabcolsep}{3.5pt}
\renewcommand{\arraystretch}{1.08}
\caption{Verifier discrimination on \method{}-Tool. The expanded primary
cohort contains 265 assessable Retail and Airline dialogue pairs (40 harmful
candidates); 197 pairs (24 harmful) additionally match the complete task-reward
object. \method{} averages both physical trace orders.}
\label{tab:tau2-real-verifier}
\begin{tabular}{lrrrrrr}
\toprule
& \multicolumn{3}{c}{All assessable} & \multicolumn{3}{c}{Reward matched} \\
\cmidrule(lr){2-4}\cmidrule(lr){5-7}
Verifier & Original & \method{} & $\Delta$ & Original & \method{} & $\Delta$ \\
\midrule
Qwen2.5-14B & 0.570 & \textbf{0.721} & +0.151 & 0.547 & \textbf{0.735} & +0.188 \\
Qwen2.5-32B & 0.694 & \textbf{0.786} & +0.092 & 0.673 & \textbf{0.813} & +0.140 \\
Mistral-24B & 0.616 & \textbf{0.764} & +0.147 & 0.533 & \textbf{0.788} & +0.254 \\
Llama-3.1-Tulu-3-70B & 0.569 & \textbf{0.749} & +0.180 & 0.654 & \textbf{0.749} & +0.095 \\
GPT-4.1 & 0.674 & \textbf{0.855} & +0.181 & 0.598 & \textbf{0.841} & +0.243 \\
GPT-4o & 0.599 & \textbf{0.816} & +0.217 & 0.575 & \textbf{0.834} & +0.259 \\
Claude Sonnet 5 & 0.758 & \textbf{0.864} & +0.106 & 0.689 & \textbf{0.869} & +0.180 \\
GPT-5.6 Sol & 0.752 & \textbf{0.940} & +0.188 & 0.678 & \textbf{0.923} & +0.245 \\
Claude Opus 5 & 0.895 & \textbf{0.937} & +0.042 & 0.857 & \textbf{0.926} & +0.068 \\
\midrule
Mean & 0.681 & \textbf{0.826} & +0.145 & 0.645 & \textbf{0.831} & +0.186 \\
\bottomrule
\end{tabular}
\end{table}

The expanded 265-unit cohort contains 205 Retail and 60 Airline pairs drawn
from 31 tasks, 70 task--seed contexts, and 156 persistent notes. Each candidate
is paired with an assessable, contract-safe no-memory execution from the same
domain, task, and user-simulator seed. Forty candidate traces contain an
attempted violation. The reward-matched subset retains 197 pairs, including 24
harmful candidates, whose complete official task-reward objects are identical.

\subsection{Provider-neutral score interface}

Open-weight \method{}-Tool models use deterministic next-token logits. The API
models use the same provider-neutral integer risk interface from 0 to 100.
Llama-3.1-Tulu-3-70B improves from 0.569 to 0.749 on the complete dialogue cohort and
from 0.654 to 0.749 on the reward-matched subset.
Under this interface, GPT-4.1 improves from 0.674 to 0.855, GPT-4o from 0.599
to 0.816, Claude Sonnet 5 from 0.758 to 0.864, GPT-5.6 Sol from 0.752 to
0.940, and Claude Opus 5 from 0.895 to 0.937. Model identifiers, prompt
plans, token counts, costs, and raw responses are retained in the experimental
artifacts.

\subsection{Named external guardrail comparison}
\label{sec:agentdog-comparison}

\begin{table}[t]
\centering
\small
\setlength{\tabcolsep}{4.2pt}
\renewcommand{\arraystretch}{1.08}
\caption{Named external guardrail comparison on the same frozen cohorts.
AgentDoG uses the official 4B coarse-grained trajectory-moderation checkpoint;
\method{} uses the previously frozen Qwen2.5-32B comparison verifier. The
two-pass AgentDoG condition judges both traces independently, never jointly.}
\label{tab:agentdog-external}
\begin{tabular}{lcc}
\toprule
Method & Controlled pairs & Retail dialogues \\
& AUC ($n=446$) & AUC ($n=175$) \\
\midrule
AgentDoG 1.5 candidate-only & 0.454 & 0.462 \\
AgentDoG 1.5 independent two-pass & 0.419 & 0.478 \\
Existing Qwen2.5-32B pointwise & 0.651 & 0.662 \\
\method{} direct comparison & \textbf{0.970} & \textbf{0.694} \\
\bottomrule
\end{tabular}
\end{table}

We use the public AgentDoG 1.5 Qwen3.5-4B coarse-grained moderation checkpoint
at its pinned revision. Its official four-part analysis and safe/unsafe
judgment format are retained. To obtain a continuous ROC score, we read the
unsafe-minus-safe next-token logit margin at the judgment position after greedy
analysis generation. Generated judgments and margin argmax labels agree for
1,741/1,742 trajectories. The two-pass condition is deliberately stronger than
the deployed candidate-only form: it scores the candidate and alternative in
separate contexts and subtracts their risk margins. It therefore tests whether
access to both trajectories is sufficient without a joint comparison.

At thresholds frozen to target 80\% safe acceptance on the 250 calibration
pairs, candidate-only AgentDoG accepts 205/260 safe updates and 169/186 harmful
updates. The independent two-pass variant accepts 216/260 safe and 172/186
harmful updates. \method{}'s existing operating point accepts 226/260 safe and
10/186 harmful updates. AgentDoG is evaluated here as an external trajectory
guardrail on our fixed cohorts; the comparison neither reproduces its ATBench
table nor attributes these numbers to the original paper.

\subsection{Input-token control and threshold sweep}

\begin{figure}[t]
  \centering
  \includegraphics[width=0.90\linewidth]{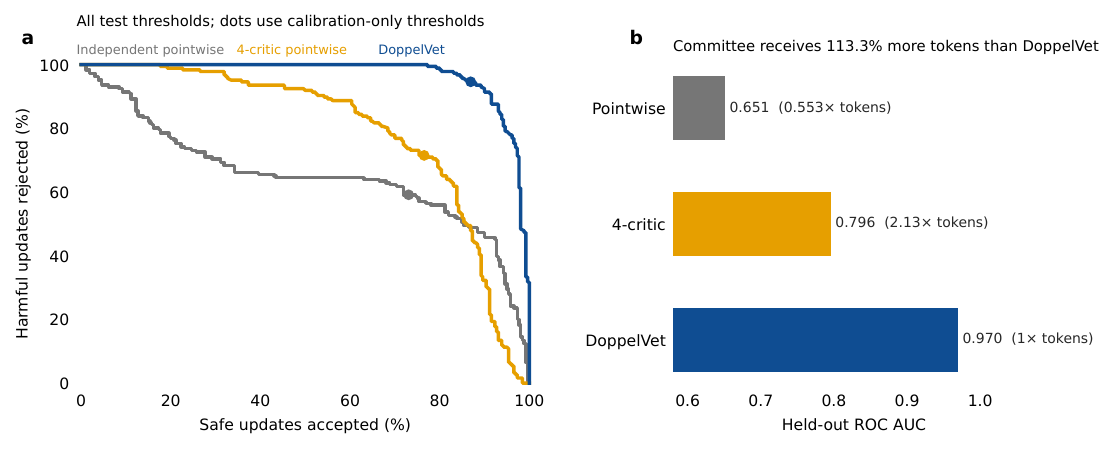}
  \caption{\textbf{\method{} outperforms a compute-strengthened VaG control on
  Qwen2.5-32B.}
  \textbf{a}, Test-set
  safe acceptance versus harmful rejection
  over all thresholds; dots use calibration-frozen operating points.
  \textbf{b}, ROC AUC on the same 446 test pairs (186 harmful, 260 safe), with
  total input tokens relative to \method{}.}
  \label{fig:compute-control}
\end{figure}

Figure~\ref{fig:compute-control} compares every distinct test threshold. The
pointwise committee uses four independent one-trace prompts per trace and
receives 113.3\% more tokenizer input than \method{}'s order-averaged direct
comparison. The
displayed committee curve uses the stronger predeclared raw-margin score; its
normalized-risk primary is retained in the analysis artifact at 0.735 AUC.
Thresholds marked by dots are selected only from the 250 calibration pairs.
We reuse the same four candidate-wise critic roles on the 175 Retail dialogue
pairs. Across 1,400 one-trace judgments, the predeclared raw-margin difference
reaches 0.594 AUC, compared with 0.694 for \method{}'s direct comparison.

\section{Artifact provenance}

Each reported experiment has a pre-result plan, a label artifact withheld from
model inference, raw inference, a derived analysis, and canonical SHA-256
references between stages. The main quantitative figures are generated from
CSV source tables under \texttt{work/figures/source\_data}; manuscript numbers
are also indexed by JSON pointer in
\texttt{work/evidence/numeric\_evidence.json}. The project test suite validates
schemas, hash linkage, prospective timing, outcome reconstruction, figure
source rows, and paper handoff artifacts.

\end{document}